\documentclass[letterpaper, 10 pt, conference]{ieeeconf}  
\IEEEoverridecommandlockouts

\usepackage{amsfonts}
\usepackage{amsmath}
\usepackage{float}
\usepackage{color}
\usepackage{tikz}
\usepackage{subcaption}
\usepackage{multicol}

 \usepackage{graphicx}  
 
\graphicspath{ {./figures/} }

\usepackage[
backend=biber,
style=numeric,
sorting=none
]{biblatex}

 \bibliography{PKDRL}

\title{\LARGE \bf
Deep Reinforcement Learning for Time Optimal Velocity Control using Prior Knowledge}

\author{Gabriel Hartmann$^{1,2}$,  Zvi Shiller$^{2}$, Amos Azaria$^{1}$
\thanks{$^{1}$Department of Computer Science, Ariel University, Israel}%
\thanks{$^{2}$Department of Mechanical Engineering and Mechatronics, 
Ariel University, Israel}
\thanks{ {\tt\small \tt\small gabriel.hartmann@msmail.ariel.ac.il, shiller@ariel.ac.il,	amos.azaria@ariel.ac.il}}%
}

\begin{document}
\maketitle
\thispagestyle{empty}
\pagestyle{empty}
\newcommand{\greenline}{\raisebox{2pt}{\tikz{\draw[-,black!40!green,solid,line width = 0.9pt](0,0) -- (5mm,0);}}}
\newcommand{\redline}{\raisebox{2pt}{\tikz{\draw[-,black!40!red,solid,line width = 0.9pt](0,0) -- (5mm,0);}}}
\newcommand{\blueline}{\raisebox{2pt}{\tikz{\draw[-,black!40!blue,solid,line width = 0.9pt](0,0) -- (5mm,0);}}}


\begin{abstract}
Autonomous navigation has recently gained great interest in the field of reinforcement learning.  However, little attention was given to  the time optimal velocity control problem, i.e. controlling a vehicle such that it travels at the maximal speed without becoming dynamically unstable (roll-over or sliding).

Time optimal velocity control can be solved numerically using existing methods that are based on optimal control and vehicle dynamics.  In this paper, we use deep reinforcement learning to generate the time optimal velocity control. Furthermore, we use the numerical solution to further improve the performance of the reinforcement learner. It is shown that the reinforcement learner outperforms the numerically derived solution, and that the hybrid approach (combining learning with the numerical solution) speeds up the training process. 
\end{abstract}



\section{Introduction}
The operation of autonomous vehicles requires the synergetic application of a few critical technologies, such as sensing, motion planning, and control.
This paper focuses on a subset of the motion planning problem, that is moving at the time optimal speeds to minimize travel time along a given path, while ensuring the vehicle's dynamic stability.  By "dynamic stability" we refer to constraints that are functions of the vehicle speed, such as rollover or sliding \cite{mann2008dynamic,altche2017simple,hwan2013optimal,petrinic2017time}.  Respecting the dynamic constraints  would thus ensure that the vehicle does not rollover or slide at any point along the path. Additional constraints that may affect the vehicle speeds, although they are not considered in this paper, include passenger comfort \cite{elbanhawi2015passenger},  traffic laws, and sensing limitations \cite{ostafew2014speed}. 
Although these constrains must be considered in most real driving scenarios, the vehicle's dynamic stability is the most challenging because it concerns the vehicle's (and passengers) safety. 




As the time optimal velocity profile is affected by the vehicle's dynamic capabilities, such as its maximum and minimum acceleration, ground/wheels interaction, terrain topography, and path geometry, a complex  dynamic model is required to ensure that the vehicle is dynamically stable during motion at any point along the path \cite{mann2008dynamic}.

Since the consideration of a detailed vehicle dynamic model may be impractical for online computation, we use a simplified model to compute the vehicle's velocity profile as discussed later. In this context, one of the goals of Reinforcement Learning (RL) is to bridge the gap between the approximate and the actual vehicle model.

A large body of work on reinforcement learning has focused on autonomous driving  with an emphasis on perception and steering \cite{bojarski2016end,michels2005high,chen2015deepdriving,drews2017aggressive,ostafew2016learning}. Some works have focused on human like velocity control \cite{zhang2018human,lefevre2016learning} or fuel  efficiency \cite{gamage2017reinforcement}.
Other works use RL to track a given reference velocity  \cite{huang2017parameterized}. 
In \cite{rosolia2018learning}, a model-predictive control is used to drive a race car at high speeds along a specific track.  The controller is tuned iteratively to reduce total motion time.  This method is applicable to repetitive tasks, where the initial state is fixed for all iterations.  Clearly, this approach is not suitable for controlling a vehicle on general paths.  We are not aware of works that use reinforcement learning of time optimal speeds along general paths, while ensuring the vehicle's dynamic stability.

This paper proposes a reinforcement learning method for driving a vehicle at the time optimal speed along a known arbitrary path.  It learns the acceleration (and deceleration) that maximizes vehicle speeds along the path, without losing its dynamic stability. 
Here, steering is not learned, but is rather determined directly by the path following controller (pure pursuit) \cite{snider2009automatic}.

One major challenge of RL is that, in many cases, the initial policy executed by the agent is random, and long training is required to achieve a good policy.  
Several methods for combining prior information about the problem into the RL process were proposed.
For example, imitation learning uses expert demonstrations (either automated or human) to train an agent in order to achieve the initial policy \cite{bojarski2016end,lefevre2016learning,zhang2018human}. The policy can then be further improved using RL \cite{sendonaris2017learning,silver2016mastering}.
In this paper we propose a different method for using prior knowledge in order to allow the RL agent to begin the training with a relatively good policy.
Instead of learning the actions directly using RL, only the variation from a nominal time optimal controller is learned by the RL agent. For this purpose, we use a numerical, model-based controller \cite{shiller1991dynamic} that controls a vehicle along a path while avoiding rollover, slipping an loosing contact with ground. This model-based method, computes a solution in a efficient way, hence it is suitable for real-time use.

The RL method, the model-based method and hybrid approach that combines both, was implemented in a simulation for a ground vehicle moving along arbitrary paths in the plane.
It is shown that, the synergy between our learning based method and the model-based method, speeds-up the learning process (especially at early stages). The RL agent that uses the model-based controller, achieves at the beginning of the learning process the same velocity as the model-based controller alone, while the pure RL approach achieves low performance at the same time. Eventually both methods converge to an average velocity that is higher by about $10\%$ than the velocity achieved by the model-based controller, while maintaining very low failure rates.

Our main contributions of this paper are (i) Applying a deep reinforcement learning-based method for driving a vehicle at time optimal speeds, subject to the vehicle's dynamic constraints, that outperforms the model-based controller; (ii) Using the  model-based prior knowledge to speed up the learning process (especially at early stages).

\section{Problem Statement} 
\label{Problem_definition}

We wish to drive a ground vehicle along a predefined path in the plane. The steering angle is controlled by a path following controller whereas its speed is determined by the learned policy. The goal of the reinforcement learning agent is to drive the vehicle at the highest speeds without causing it to rollover or  deviate from the defined path beyond a predefined limit.   

\begin{figure}[h]
  \centering
  \includegraphics[width=\linewidth]{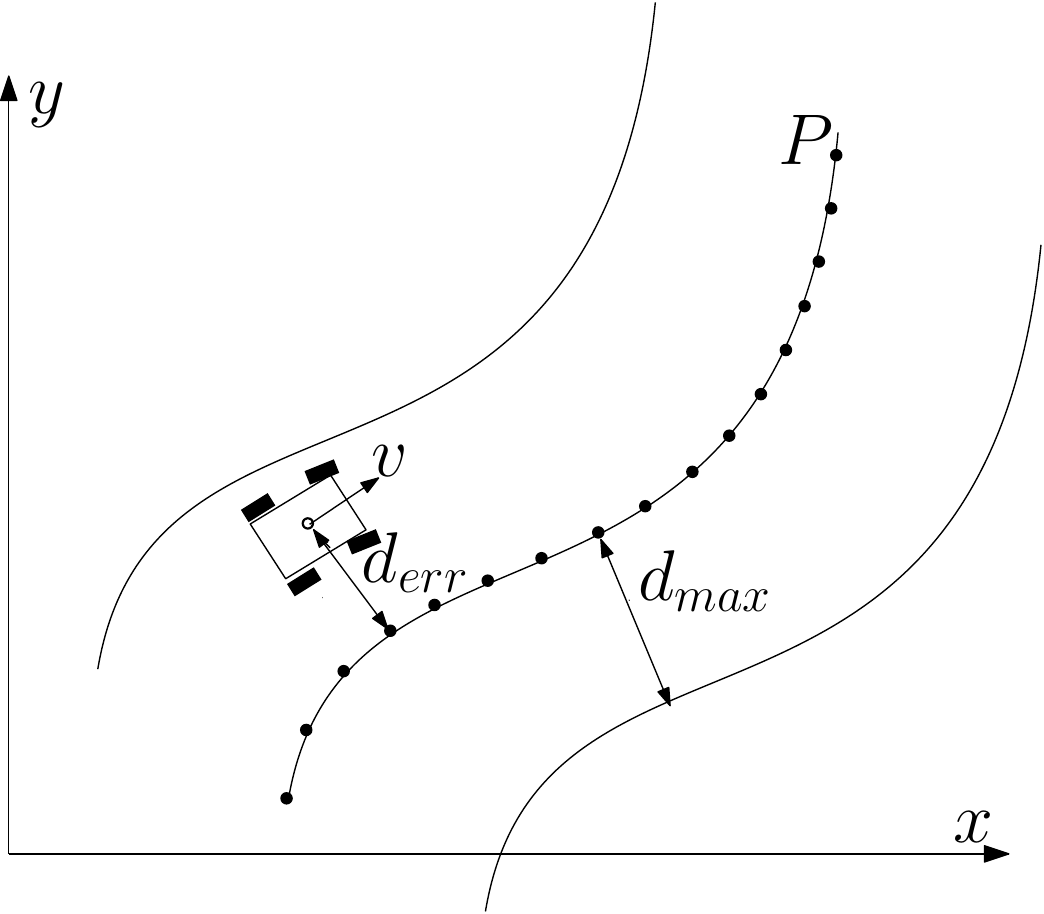}
  \caption{A vehicle, tracking path $P$ within the allowed margin $d_{max}$.} 
  \label{fig:definitions}
\end{figure}

The path is defined by $P$, $P=\{{{p}_{1}},{{p}_{2}},\cdots,{{p}_{N}}\}$,  ${{p}_{i}}\in \mathbb{R}^2$, $i \in \{1,2,\cdots,N\}$.
The position of the vehicle's center of mass is denoted by $ q \in \mathbb{R}^2 $,  yaw angle  $\theta$, and roll angle $\alpha$. The vehicle's  speed is $v \in \mathbb{R}$, $0\le v \le {{v}_{\max }}$.
The throttle (and brakes) command that affects the vehicle's acceleration (and deceleration)  is $\tau \in [-1,1]$. 
The steering control of the vehicle is performed by a path following controller (pure pursuit \cite{snider2009automatic}).  The deviation of the vehicle center from the desired path is denote by $d_{err}$, as shown in Fig. \ref{fig:definitions}.   


The agent's goal is to 
drive the vehicle at the maximal speed along the path, without losing its dynamic stability (sliding and rollover), while staying within a set deviation from the desired path, i.e.  $d_{err} \le d_{\max }$, and within a "stable" roll angle, i.e.  $|\alpha| \le \alpha _{\max }$, where ${\alpha }_{\max }$ is the maximal roll angle beyond which the vehicle is statically unstable.




%
The time optimal policy maximizes the speed along the path during a fixed distance.
More formally, for every path $P$ with length $D$, and a vehicle at some initial velocity ${v}_{init}$, initial position $q$, which is closest to point $p_i\in P$ along the path, we wish to derive the 
time optimal policy $\pi^*$ that at every time $t$ outputs the action $\tau = \pi^*(s_t)$ that maximizes the vehicle speed (minimizing traveling time), while  
ensuring that every state $s_{t}$ is stable. 
%
%
The time optimal velocity along path $P$ is the velocity profile $v(t)$ produced by the optimal policy $\pi^*$.

\section{Time Optimal Velocity Control using Reinforcement Learning}
\label{sec:Learning_method}
Our basic reinforcement learner is a direct adaptation of the ``Deep Deterministic Policy Gradient'' (DDPG) \cite{lillicrap2015continuous} to the time optimal velocity  control  problem. We refer to this method as the Reinforcement learning based Velocity Optimizer REVO. 

\subsection{Deep Deterministic Policy Gradient}
  DDPG \cite{lillicrap2015continuous} is an actor-critic, model-free algorithm for a continuous action space, $A$, and a continuous state space, $S$. The agent is assumed to receive a reward ${{r}_{t}}\in \mathbb{R}$ when being at state ${{s}_{t}}\in {{\mathbb{R}}^{\left| S \right|}}$ and taking action ${{a}_{t}}\in {{\mathbb{R}}^{\left| A \right|}}$. The transition function  $p({{s}_{t+1}}|{{s}_{t}},{{a}_{t}})$ is defined as the probability of ending at ${{s}_{t+1}}$ when being at state ${{s}_{t}}$ and taking action ${{a}_{t}}$. The goal of the DDPG algorithm is to learn a deterministic policy $\pi :S\to A$ (represented as a neural network) that maximizes the return from the beginning of the episode: \[{{R}_{0}}=\sum\limits_{i=1}^{T}{{{\gamma }^{(i-1)}}r({{s}_{i}},\pi ({{s}_{i}}))}\] where $\gamma \in [0,1]$ is the discount factor. DDPG learns the policy using policy gradient. The exploration of the environment is done by adding exploration noise to the actions.

We use DDPG to train an agent for driving along any given path at the highest possible speed, while preventing a rollover or slipping away from the path.
The training process consists of episodes; at each episode the vehicle moves along a randomly generated path. Training an agent on randomly generated paths allow the learned policy to be more general. Only paths that are kinematically feasible are considered, that is, the generated paths do not contain any sharp curves that exceed the vehicle's minimum turning radius (the maximum turning ability of the vehicle at zero velocity). 
Each path, $P$, is generated by smoothly connecting short path segments of random length and curvature until reaching the desired length. 
This ensures that the selected path respects the vehicles steering capabilities. 

The state, $s$, includes a down-sampled limited horizon path segment, ${{P}_{s}}\subseteq P$, which is defined relative to the vehicle's position and the vehicle speed, $v$.  More formally, \[P_s =\{{{p}_{m}},p_{m+d},p_{m+2d}\cdots {{p}_{m+kd}}\}\] 
where $m$ is the index of the closest point on the path $P$, to the vehicle, $d\in \mathbb{N}$ is the down-sampling factor and $k\in \mathbb{N}$ is a predefined number of points. 
In addition to this path segment, also the current velocity of vehicle ($v$) is included in the state. 
Therefore, the state of the system is defined as $s=\{v, P_{s}\}$. 

The DDPG agent is not provided with any information related to the path segment following $p_s$. Therefore, $p_s$ is required  to be long enough in order to enable the vehicle to decelerate to a safe velocity at the end of this path segment, even when driving at the maximal speed. If $p_s$ is too short, the agent may need to drive at a lower speed to prepare for any unforeseen curve that might appear as the vehicle moves forward.   


The reward function is defined as follows:
If the vehicle is stable and has a positive velocity, the reward $r$ is proportional to the vehicle's velocity ($r_t = k v_t, k \in \mathbb{R}_+ $). If the vehicle encounters an unstable state, it receives a negative reward. To encourage the agent not to stop the vehicle during motion, a small negative reward is received if $v_t=0$.

At each time step, the action is determined as ${{a}_{t}}={{\tau}_{t}}=\pi ({{s}_{t}})+\eta(t)$ where $\eta(t)$ is the exploration noise. 
The episode terminates at time $T$ or if the vehicle becomes unstable.


\section{Computing the Time optimal Velocity Profile}
The time optimal velocity profile of a vehicle moving along a specified path can be numerically computed  using an efficient algorithm described in \cite{shiller1991dynamic,z.shillerh.lu1992,mann2008dynamic}.  It uses optimal control to compute the fastest velocity profile along the given path, taking into account the vehicle's dynamic and kinematic models, terrain characteristics, and a set of dynamic constraints that must be observed during the vehicle motion: no slipping, no rollover and maintaining contact with the ground at all points along the specified path. This algorithm is used here as a model predictive controller, generating the desired speeds at every point along a path segment ahead of the vehicle's current position. 
This Velocity Optimization using Direct computation is henceforth termed VOD.  
The output of this controller is used to evaluate the results of the learning based optimization (REVO), and to serve as a baseline for the training process. We now briefly describe the algorithm in some details.  

Given a vehicle that is moving along a given path $P$, the aforementioned algorithm  computes the time optimal velocity, under the following assumptions:
\begin{itemize}
\item The dynamics of the vehicle are deterministic; 
\item The vehicle moves exactly on the specified path i.e. $(p_{err})_t = 0 $, $t=\{0,...N\}$;
\item The vehicle is modeled as a rigid body (no suspension); 
\item Vehicle parameters, such as geometric dimensions, mass, the maximum torque at the wheels, the coefficient of friction between the wheels and ground, are known.
\end{itemize}

These assumptions help  simplify the computation of the time optimal velocity profile.  This simplification does not seriously affect our approach as the goal of the learning process is to bridge the gap between the model and reality, which may always exist, regardless of the  fidelity of the theoretical model.

The algorithm first computes the maximal velocity profile along the path, termed the "velocity limit curve", which represents the highest vehicle speeds, above which at least one of the vehicle's dynamic constraints is violated, i.e. the vehicle either rolls-over, slides, or looses contact with the ground.     
The velocity limit is determined by the coefficient of friction between the wheels and ground as well as by the centripetal forces that might cause the vehicle to slide or rollover.

The time optimal velocity profile is computed by applying ``Bang-Bang'' acceleration, i.e. either maximum or minimum acceleration, at all points along the path.  Bang-bang control is known to produce the time optimal motion of second order systems \cite{bryson1969applied}. The optimal velocity profile is computed by integrating forward and backwards the extreme accelerations at every point along the path so as to avoid crossing the velocity limit curve \cite{z.shillerh.lu1992}.

\begin{figure}[h]
\centering
\begin{subfigure}{\linewidth}
  \centering
  \includegraphics[width=\linewidth]{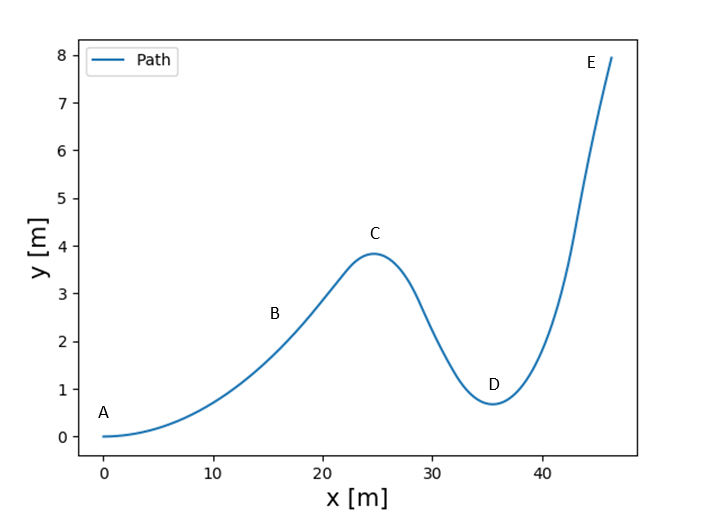}
  \caption{}
  \label{fig:simple_path}
\end{subfigure}%

\begin{subfigure}{\linewidth}
  \centering
  \includegraphics[width=\linewidth]{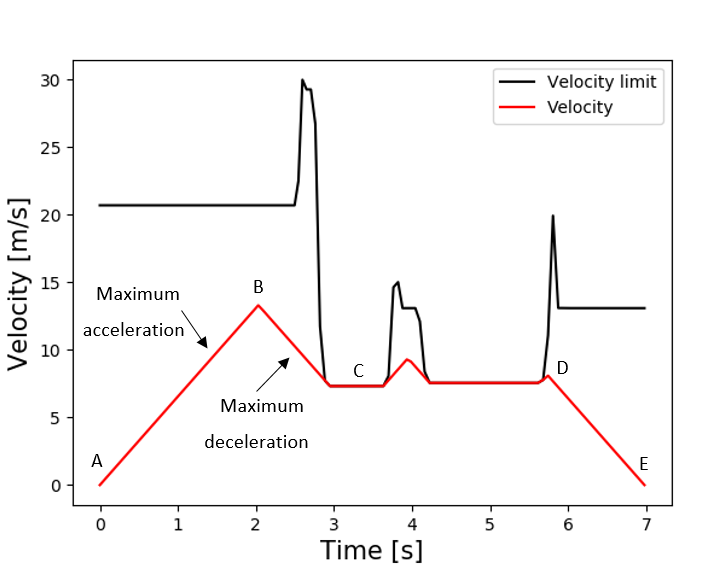}
  \caption{}
  \label{fig:simple_path_vel}
\end{subfigure}
\caption{(a) A curved path segment (b) The directly computed optimal velocity profile (red) and the velocity limit curve (black). The velocity limit drops along sharp curves along the path. The optimal velocity never crosses the velocity limit curve. }
\label{fig:simple_path_figure}
\end{figure}

Fig. \ref{fig:simple_path} shows a given planar curved path. The velocity limit curve along that path is shown in black in Fig. \ref{fig:simple_path_vel}. Note the drops in the velocity limit caused by the sharp curves $C$ and $D$ along the path. Clearly, moving at high speeds along these curves might cause the vehicle to either slide or rollover (which of the two occurs first, depends on the location of the vehicle's center of mass).    
The optimal velocity thus starts at zero (the initial boundary condition), accelerates at a constant acceleration until point $B$, where it decelerates to avoid crossing the velocity limit towards point c.  At point $D$, the optimal velocity decelerates to a stop at the end point $E$ (the assumed final  boundary condition).  


The velocity computed by this algorithm is used to control the vehicle along the specified path. At every time $t$, the optimal velocity profile is computed along the limited horizon path segment $P_s$ (as was formally defined in Section \ref{sec:Learning_method}). The vehicle's speed at time $t$ serves as the initial condition for the velocity profile computed from that point.  To ensure that the vehicle can decelerate to a stop at the end of this path segment, the target velocity at the endpoint of $P_s$ is set to zero. The action produced by the controller at time $t$ is the initial acceleration of the velocity profile computed at time $t$. This acceleration is used as a command to the vehicle's engine. This controller is used as a baseline for REVO.

\section{Using Direct Computation to Enhance Reinforcement Learning}

\begin{figure}[h]
\centering
\begin{subfigure}{\linewidth}
  \centering
  \includegraphics[width=.5\linewidth]{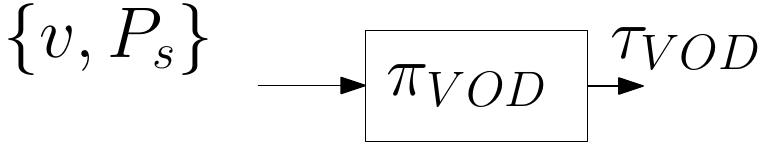}
  \caption{}
 \label{fig:VOD}
\end{subfigure}
\vspace{0.05\linewidth}%

\begin{subfigure}{\linewidth}
  \centering
  \includegraphics[width=.52\linewidth]{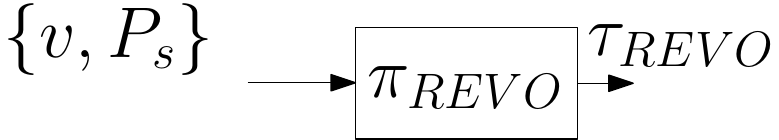}
  \caption{}
  \label{fig:REVO}
\end{subfigure}
\vspace{0.05\linewidth}%

\begin{subfigure}{\linewidth}
  \centering
  \includegraphics[width=.65\linewidth]{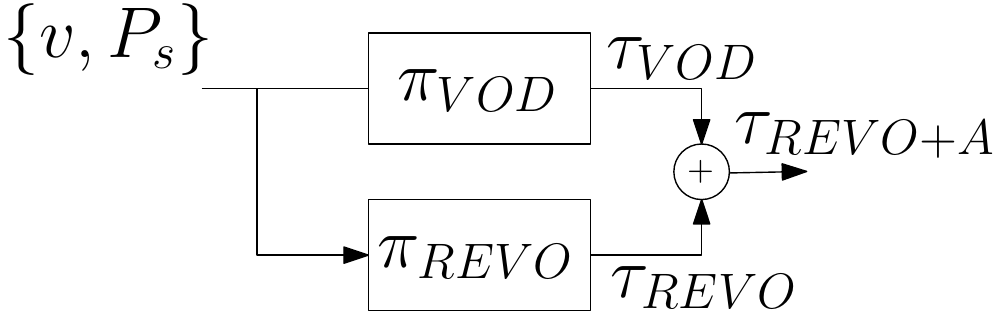}
  \caption{}
  \label{fig:REVO+A}
\end{subfigure}%
\vspace{0.05\linewidth}%

\begin{subfigure}{\linewidth}
  \centering
  \includegraphics[width=.78\linewidth]{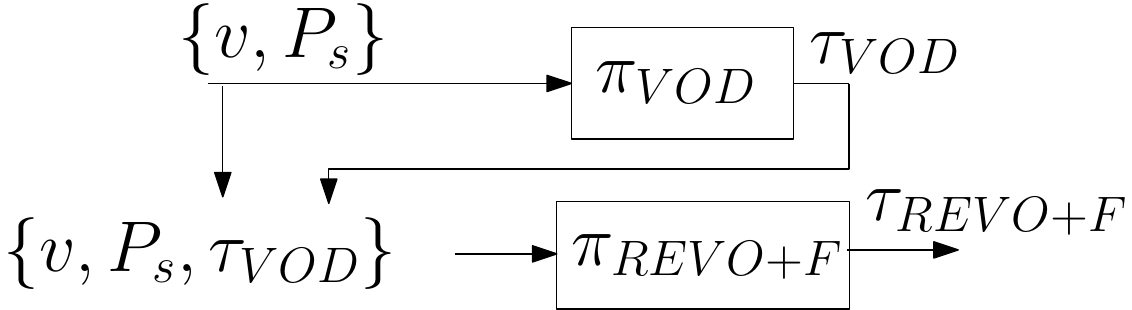}
  \caption{}
  \label{fig:REVO+F}
\end{subfigure}


\caption{ $\pi$ is a policy, $\tau$ is an action, $\{v,P_s\}$ is the state. (a) VOD: Direct planning (b) REVO: DDPG based learning. (c) REVO + A: combines the actions of VOD with REVO. (d) REVO + F: adds the action output of VOD as a feature in the state space of REVO.}
\label{fig:methods}
\end{figure}

In this paper, we propose to speed-up the learning process by
combining VOD (the direct velocity optimization controller) with REVO (the reinforcement learning based controller). This is done by first adding the actions $\tau_{VOD}$ and $\tau_{REVO}$ of VOD and REVO, respectively, to produce the action $\tau _{REVO+A}$ of the combined policy  REVO+A (REVO+Action):   
\[{\tau }_{REVO+A}={{\tau }_{VOD}}+{{\tau }_{REVO}}.\] 
The REVO+A policy is illustrated in Fig. \ref{fig:REVO+A}.

The REVO+A policy first follows the actions of the VOD controller, i.e.
${\tau_{REVO+A}}\approx {\tau_{VOD}}$ because ${\tau _{REVO}} \approx 0$ at the beginning of the learning process.  This is significantly better than a randomly initialized policy as in $\pi_{REVO}$. 
It simplifies the problem for the reinforcement learner agent, which only learns the deviation from VOD, as oppose to learning the actions from ground up.

The second approach proposed in this paper to combining REVO and VOD is based on adding the action output $\tau_{VOD}$ from the VOD controller as an additional feature to the state space of the agent:  \[s=\{{{\tau }_{VOD}}, {v}, {{P}_{s}}\}.\] 
We denote this method REVO+F (REVO+Feature). It is illustrated in Fig. \ref{fig:REVO+F}. 

An intuitive justification for using REVO+F  is that the reinforcement learner has the information about ${{\tau }_{VOD}}$, and hence, the agent can use this information to improve its actions.



\section{Experimental Results}
\label{sec:result}

The performance of the proposed methods were tested in several experiments as detailed henceforth.    

\subsection{Settings}
A simulation of a four-wheel vehicle was developed using ``Unity'' software \cite{unity}. 
A video of the vehicle driving along a path at time optimal velocity is available at \cite{video}.

The vehicle properties were set to  width= $2.1 m$,  height=$1.9 m$,  length=$ 5.1 m$,  center of mass at the height of $0.9 m$,  mass=$3,200 Kg$, and a total force produced by all wheels of $21 KN$. 

    
   
    
    

The maximal velocity of the vehicle was set to $v_{max}= 30 m/s$ ($108 km/h$). 
Note that the actual speed limit is determined by the path, which may be lower in most cases than the above set limit.    

The maximal acceleration of the vehicle is $6.5 m/s^2$. The acceleration and deceleration are applied to all four wheels (4x4); steering is done by the front wheels (Ackermann steering). The  friction coefficient between the wheels and the ground was set arbitrarily high (at 5) to focus this experiment on rollover only.


Each episode is limited to $100$ time steps. The time step is set to 0.2 seconds, i.e. $20$ seconds per episode. The policy updates are synchronized with the simulation time steps, two updates per step. $P_s$ consists of 25 points along the path ahead of the vehicle ($\left| P_{s} \right|=25$. The distance between one point to the next point in $P_s$ is $1 m$. \[{|{p}_{i}}-{p}_{i+1}|=1[m]: {{p}_{i}},{{p}_{i+1}}\in P_{s}, i\in \{0,1,\cdots ,25\}\]
A state is considered unstable if the roll angle of the vehicle exceeds 4 degrees (${\alpha }_{\max } = 4$), and when the vehicle deviates more then $2 m$ ($d_{max} = 2$) from the nominal path.
The reward function was defined as:
\[\left\{ \begin{matrix}
   -1  \\
    {0.2 v/v_{max}} \\
   -0.2  \\
\end{matrix}\begin{matrix}
   \ \ \  s\,\,\mbox{is not stable}  \\
   \  s\,\, \mbox{is stable}  \\
   v = 0  \\
\end{matrix} \right.\]  

All the hyper-parameters of the reinforcement learning algorithm (e.g. neural network architecture, learning rates)  were set as described in \cite{lillicrap2015continuous}.

\subsection{Experiment Protocol}
During the training process, the vehicle drives along randomly generated paths using the learned policy with exploration noise.
Each training process is performed until reaching 90,000 policy updates. 
Every  $5000$ updates the neural networks parameters are saved for evaluation.
To evaluate the policy during the training process, the vehicle runs along $100$ random paths on every saved parameter set. During the evaluation, the exploration noise is disabled.
This training and evaluation process was repeated 5 times for each of the methods. 

 
The agent's goal is to maximize its average velocity. 
Since the average velocity during the failed episodes was usually higher than the average velocity during successful episodes, we excluded failed episodes when presenting the average velocity of each method. 


\subsection{Results}
\begin{figure}[h]
\centering
\begin{subfigure}{\linewidth}
  \centering
  \includegraphics[width=\linewidth]{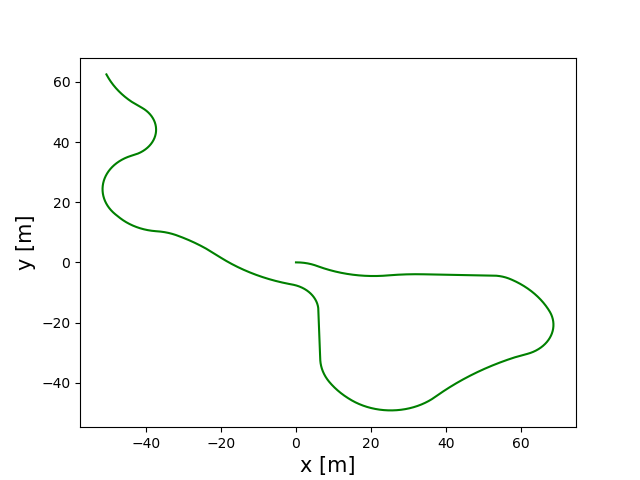}
  \caption{}
  \label{fig:random path}
\end{subfigure}%

\begin{subfigure}{\linewidth}
  \centering
  \includegraphics[width=\linewidth]{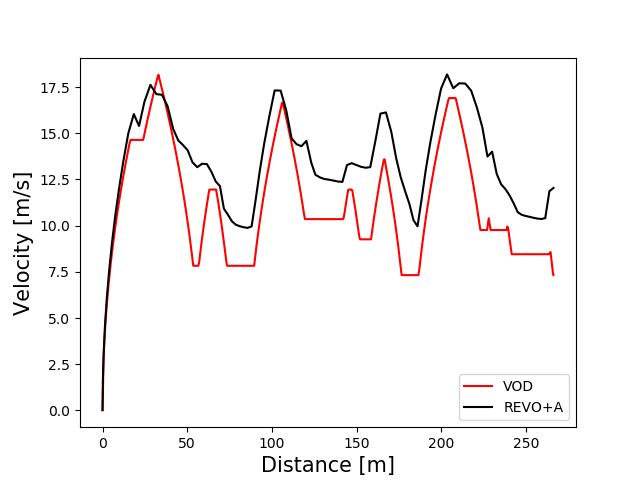}
  \caption{}
  \label{fig:vel_lim}
\end{subfigure}
\caption{(a) Example of a random path  (b) The dynamics based velocity profile (VOD) and the velocity profile of a trained REVO+A agent.}
\label{fig:example path}
\end{figure}

Fig. \ref{fig:random path} shows an example path, and Fig. \ref{fig:vel_lim} shows the velocity profile along this path during $20$ seconds, for both the VOD controller (red) and REVO+A after convergence (black). As can be seen, the learned velocity profile of REVO+A is higher than that of the VOD.

 \begin{figure}[h]

  \centering
  \includegraphics[width=.9\linewidth]{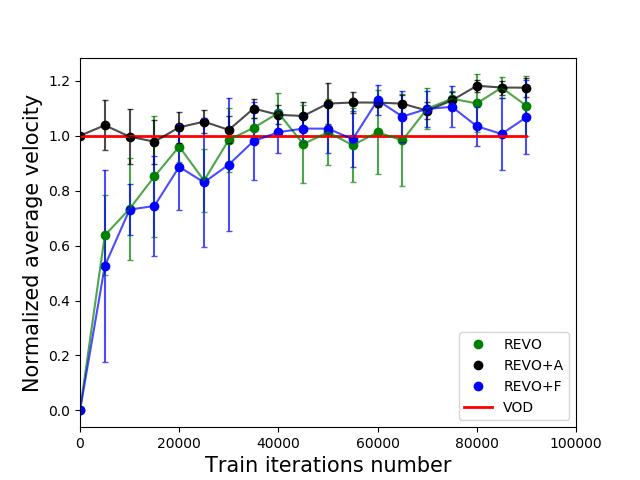}
\caption{Average velocity on $100$ random paths, measured at every $5000$ training steps (normalized with respect to VOD), on $5$ different training processes. The bars represent the variance between the training processes.}
  \label{fig:progress}
\end{figure}

\begin{figure}[h]
  \centering
  \includegraphics[width=\linewidth]{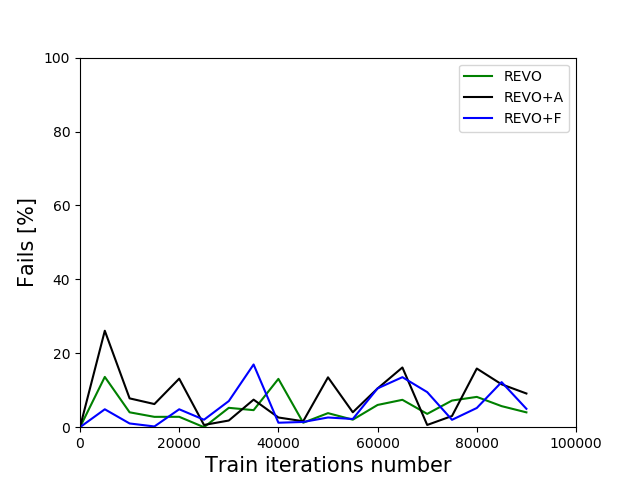}
    \caption{The failure rate of all methods during training.}
    \label{fig:fails}
\end{figure}

Fig. \ref{fig:progress} presents the normalized average velocity along the path during each episode. 
All  results were normalized with respect to the VOD, hence it appears as  a horizontal line at 1.0.

At the beginning of the training process, REVO did not achieve any progress;  after about $40,000$ training updates, REVO achieved the same performance as that of VOD. 
REVO+A achieved the same performance as VOD from the very beginning. This implies that REVO+A converges much faster than REVO, because REVO+A uses VOD as a baseline.

When the training process continues, the policies learned by all methods improve the performance of the vehicle's velocity compared to using VOD by about $10\%$. This is expected because VOD uses a relatively simple vehicle model.  


REVO+F doesn't improve  the converge time compared to REVO, in this experiment.  On the other hand,  REVO+F performed better than REVO when used in a different setting, as was shown in Section \ref{sec:closer_look}.

    
   
    
    

The failure rate of the different methods has a relatively high variance as is depict in Fig. \ref{fig:fails}. After training and evaluation, it is possible to choose the best policy that achieves high velocity and low failure rates. 
When re-evaluating the best policies achieved by all methods on $1000$ new episodes, the failure rate is lower than $1\%$ and the average velocity is approved to be statistically significant higher that VOD by about $10\%$ (using student's t-test, $p<0.0001$).

    
   
    
    


\subsection{Near Optimality of VOD}

VOD uses a computational effective model to compute the velocity. 
In this section we show that the VOD velocity cannot be easily increased without resulting in high failure rates.
We show that even slightly scaling up the velocity of the VOD policy, causes the vehicle to fail. This implies that the velocity computed by VOD is close to the real performance envelope.

\begin{figure}[h]
  \centering
  \includegraphics[width=\linewidth]{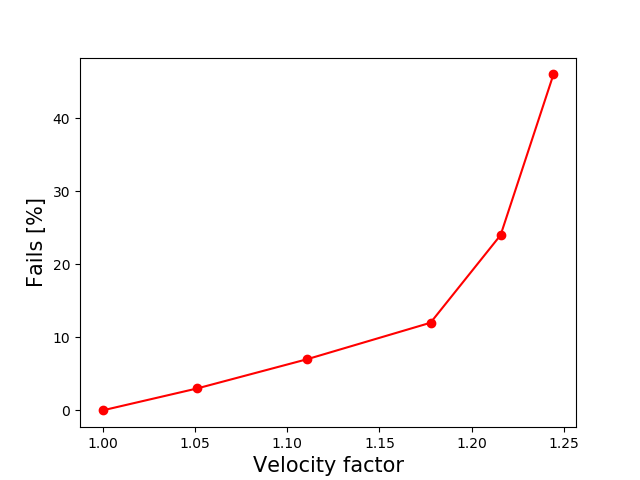}
  \caption{Failure rate of VOD when scaling up the VOD velocity} 
  \label{fig:fails_velocity_factor}
\end{figure}

Fig. \ref{fig:fails_velocity_factor} shows, that scaling up the VOD velocity, cause an increased failure rate (evaluated on $100$ episodes at 6 different velocity factors between $1.00$ and $1.25$). As depicted by the figure, when the velocity is scaled up by $5\%$, the vehicle fails on $3\%$ of the episodes, and scaling up by $20\%$ results in a failure rate of nearly $50\%$.

When controlling the vehicle using the trained policies of REVO, REVO+A and REVO+F,
a higher velocity (by about $10\%$ can be achieved without increasing the failure rate. 

\subsection{Closer Look at REVO+F}
\label{sec:closer_look}
\begin{figure*}[h]
  \begin{subfigure}{\textwidth}
  \includegraphics[width=.25\linewidth]{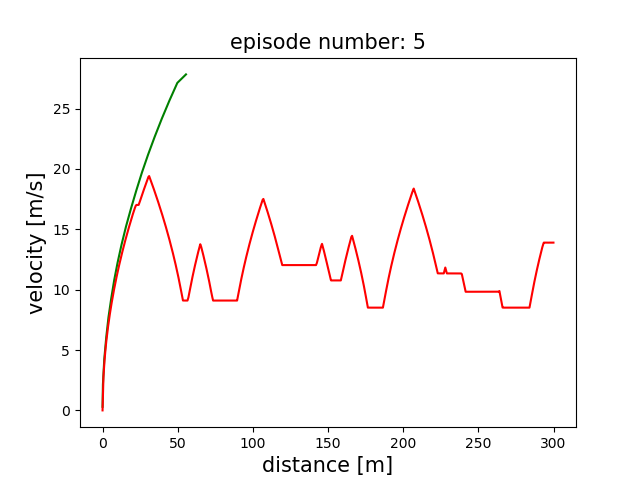}\hfill
  \includegraphics[width=.25\linewidth]{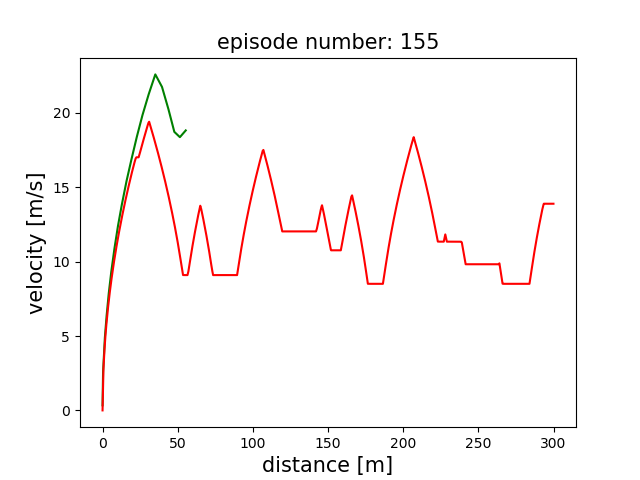}\hfill
  \includegraphics[width=.25\linewidth]{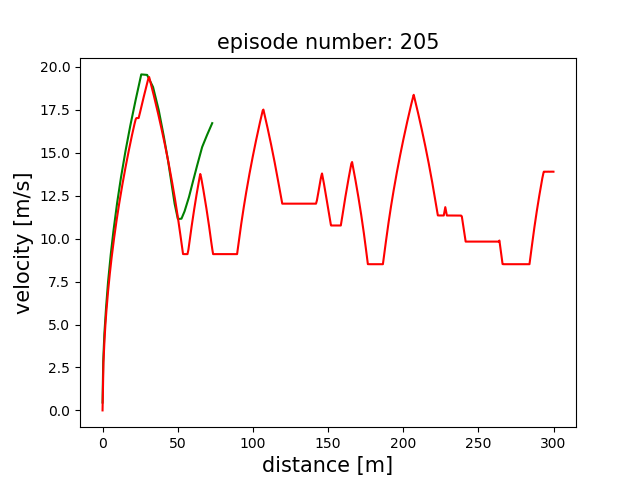}\hfill
  \includegraphics[width=.25\linewidth]{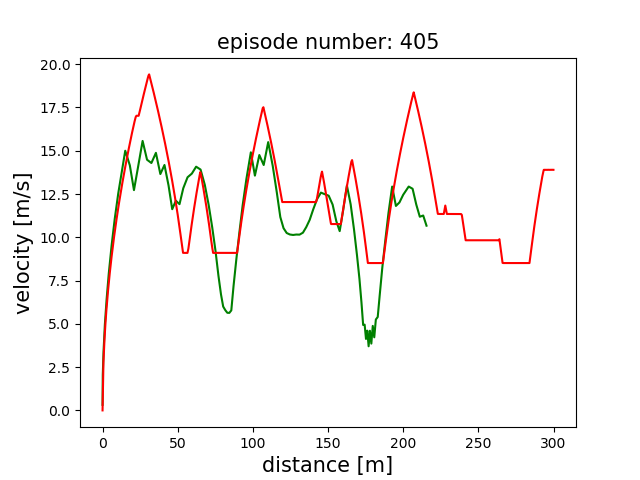}
  \caption{\protect\greenline REVO, \ \ \protect\redline VOD}
  \end{subfigure}\par\medskip
  \begin{subfigure}{\textwidth}
  \includegraphics[width=.25\linewidth]{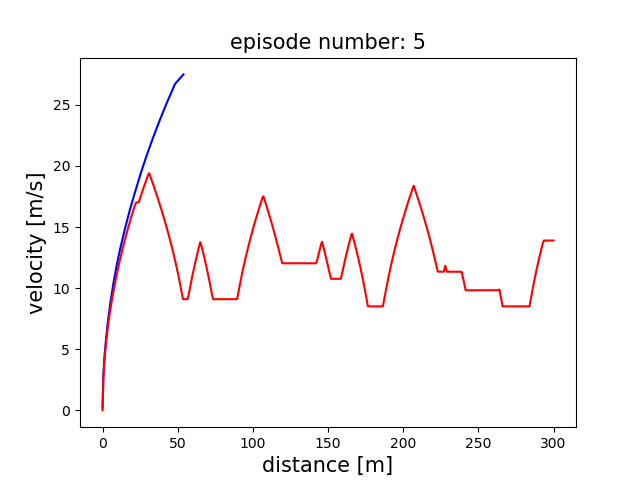}\hfill
  \includegraphics[width=.25\linewidth]{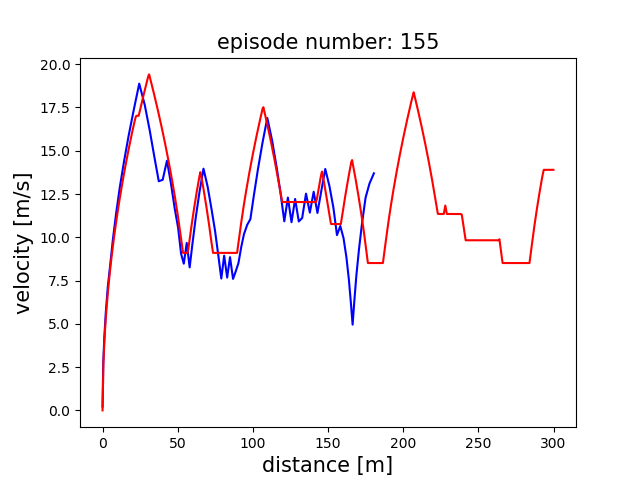}\hfill
  \includegraphics[width=.25\linewidth]{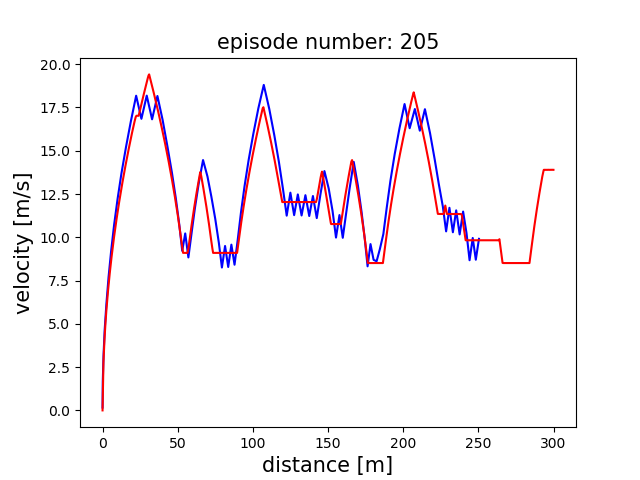}\hfill
  \includegraphics[width=.25\linewidth]{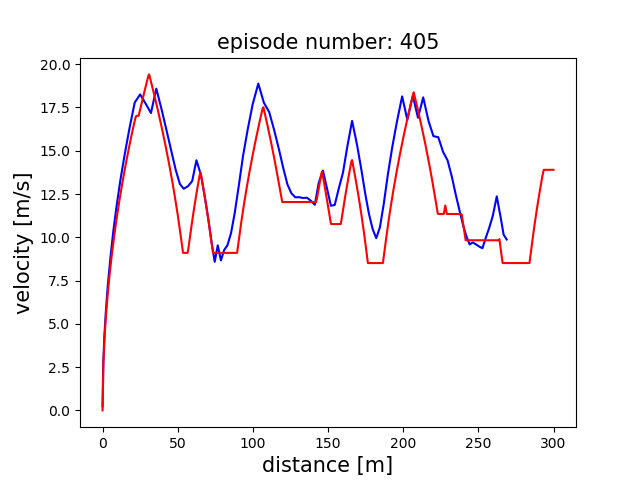}
  \caption{\protect\blueline REVO+F, \ \  \protect\redline VOD}
  \end{subfigure}\par\medskip

  \caption{A comparison between learning progress of REVO and REVO+F. In episode 5, both methods accelerate until a roll-over occurs. In episode 155, REVO+F started to imitate the VOD controller, while REVO shows a very little progress. In episode 205, D-VOL shows almost full imitation of VOD velocity, while REVO  is still in its initial stages of learning. In episode 405, REVO+F actions result in a higher velocity than VOD }
  \label{fig:feature influence}
\end{figure*}

Before we conclude this section, we would like to take a closer look at how adding the VOD output ($\tau_{VOD}$) as a feature to the state (REVO+F) influences the training process. When running the training process on a single randomly picked path (instead of training each episode on a new path) it is possible to closely track the policy improvement. In this case, as can be seen in Fig. \ref{fig:feature influence}, after some training, the learned policy uses the VOD information supplied through the additional feature, hence the velocity profile is similar to that of VOD; while the policy achieved by the regular training process (REVO) is still not able to complete this path. More research is required to understand this observation better.

\section{Conclusions}
\label{sec:conclusion}

 In this paper, we addressed the issue of deep reinforcement learning of autonomous driving at high speeds along specified paths, while accounting for the vehicle dynamics and its dynamic constraints (rollover and sliding).  
 To this end, we proposed two methods, each combine traditional deep reinforcement learning (REVO) with a direct computation of the time optimal velocity profile along a given path (VOD). One method, denoted REVO+A,  adds actions of REVO and VOD so that it is initialized at the VOD profile, and thus it learns only the required deviations from the model-based optimal speeds. The second method, denoted REVO+F, adds the action of VOD as a feature to the state of REVO.  
 
 The two methods were tested in experiments using a simulator that simulates the dynamics of a real vehicle.   
 We show that REVO+A results in a significant improvement to the basic reinforcement learner REVO, especially at early stages of the learning process.  It was shown that the REVO took around $40,000$ iterations to converge to an model-based velocity controller (VOD), compared to  an immediate convergence by the combined controller (REVO+A).
 Another interesting result was that the learning process improved over the model-based velocity profile. This is not surprising as we used a relatively simple and computational effective  vehicle model to speed up computation and the learning process. 
 
 The REVO+F method showed no significant advantage for randomly chosen paths.  However, when learning to drive along a single path, it quickly converged to the VOD velocity profile.  This suggests that the REVO+F agent quickly recognizes the utility of the model-based velocity profile. Further research may be required in order to take advantage of this phenomenon.      







\printbibliography[heading=bibintoc]
\end{document}